\documentclass[10pt, a4paper]{article}

\usepackage[final]{lrec2026}

\usepackage{float}
\usepackage{amssymb}
\usepackage{tikz}
\usetikzlibrary{arrows.meta, positioning, calc, backgrounds}
\usepackage{algorithm}
\usepackage{algpseudocode}
\usepackage{booktabs}
\usepackage{enumitem}
\usepackage{multirow}
\usepackage{amsmath}
\usepackage{longtable,array}
\usepackage{supertabular}

\usepackage{listings}
\usepackage{xcolor}

\lstdefinestyle{prompt}{
  basicstyle=\ttfamily\scriptsize,
  breaklines=true,
  breakatwhitespace=true,
  breakautoindent=true,
  frame=single,
  framesep=4pt,
  xleftmargin=2pt,
  xrightmargin=2pt,
  aboveskip=6pt,
  belowskip=6pt,
  columns=fullflexible,
  keepspaces=true,
  showstringspaces=false,
}

\title{Retrieving Climate Change Disinformation by Narrative}

\name{\parbox{\textwidth}{\centering\textbf{Max Upravitelev$^{1,2}$, Veronika Solopova$^{1,2}$, Charlott Jakob$^{1,2}$, \\[-3pt]
Premtim Sahitaj$^{1,2}$, Sebastian Möller$^{1,2}$ and Vera Schmitt$^{1,2,3,4}$}}}

\address{$^{1}$Technische Universit\"{a}t Berlin, 
         $^{2}$German Research Center for Artificial Intelligence (DFKI) \\
         $^{3}$BIFOLD -- Berlin Institute for the Foundations of Learning and Data \\
         $^{4}$Centre for European Research in Trusted AI (CERTAIN) \\
         max.upravitelev@tu-berlin.de}

\abstract{
Detecting climate disinformation narratives typically relies on fixed taxonomies, which do not accommodate emerging narratives. Thus, we re-frame narrative detection as a retrieval task: given a narrative's core message as a query, rank texts from a corpus by alignment with that narrative. This formulation requires no predefined label set and can accommodate emerging narratives. We repurpose three climate disinformation datasets (CARDS, Climate Obstruction, climate change subset of PolyNarrative) for retrieval evaluation and propose SpecFi, a framework that generates hypothetical documents to bridge the gap between abstract narrative descriptions and their concrete textual instantiations. SpecFi uses community summaries from graph-based community detection as few-shot examples for generation, achieving a MAP of 0.494 on CARDS without access to narrative labels. We further introduce narrative variance, an embedding-based difficulty metric, and show via partial correlation analysis that standard retrieval degrades on high-variance narratives (BM25 loses 63.4\% of MAP), while SpecFi-CS remains robust (32.7\% loss). Our analysis also reveals that unsupervised community summaries converge on descriptions close to expert-crafted taxonomies, suggesting that graph-based methods can surface narrative structure from unlabeled text.
\\ \newline 
}

\begin{document}

\maketitleabstract

\section{Introduction}

Recent datasets on climate change disinformation \citeplanguageresource{lr-CARDS}, \citeplanguageresource{lr-PN}, \citeplanguageresource{lr-CO} organize individual claims under narrative taxonomies defined by core messages. These taxonomies group together texts, sometimes with little lexical overlap: the claim ``\textit{Carbon dioxide is vital to all life on Earth because no vegetation can exist without it}'' and a lengthy scientific rebuttal arguing that ``\textit{the historical increase in the atmosphere's CO2 concentration has been good for the Amazon's trees}'' share near-zero Jaccard similarity, yet both serve the same narrative: that CO$_2$ is essentially plant food (a narrative from the CARDS taxonomy \citeplanguageresource{lr-CARDS}).

Framing narrative identification as classification enables the detection of known narratives but limits adaptability: classification assumes a fixed label set, whereas disinformation narratives evolve. Re-framing the task as retrieval, where a narrative's core message serves as a query to rank candidate texts, enables a more flexible monitoring strategy that can target emerging, previously unseen narratives. In practice, this means that when fact-checkers or journalists observe a potentially emerging narrative, they could formulate its core message as a query and search a corpus to assess how prevalent it already is without requiring a predefined label set or retraining a classifier. 
However, this flexibility comes at a cost: narrative retrieval poses its own challenges. Unlike standard semantic search, which matches surface-level meaning, narrative retrieval must identify texts by their underlying core message, which may never be stated explicitly. Prior work has shown that dense retrievers fail to respect implicit logical constraints in queries \citep{shen2025logicol} and that text embedding models struggle with structural and relational understanding between concepts \citep{2023syntax_well}. Narrative understanding specifically remains a known limitation of current language models \citep{zhu-etal-2023-nlp}. These failures cascade in narrative retrieval, where queries express abstract core messages (e.g., ``CO$_2$ is plant food'') that texts may support through implicit logical entailment or varied syntactic framings without stating the theme directly: the difficulty is the gap between narrative descriptions, which are abstract, and their textual instantiations, which are concrete and stylistically diverse.

In this paper, we explore narrative retrieval in the domain of climate disinformation. Our primary contributions are analytical rather than architectural: the individual components of our pipeline, including dense retrieval, dynamic few-shot sampling, HyDE-style generation \citep{2023hyde}, and graph-based community detection via the framework NodeRAG \citep{2025noderag}, are drawn from existing work. Their combination serves as the experimental setup for three contributions:

\begin{enumerate}[leftmargin=*, nosep]

\item \textbf{Retrieval-based evaluation of narrative datasets.} We repurpose three climate disinformation narrative datasets (CARDS, Climate Obstruction, a climate change-related subset of PolyNarrative) for retrieval evaluation, using narrative labels as queries and associated texts as relevance judgments.

\item \textbf{SpecFi: Speculative Fiction generation for narrative retrieval.} We propose a framework that bridges the gap between abstract narrative descriptions and concrete textual instantiations by generating hypothetical documents following the HyDE strategy \citep{2023hyde}. SpecFi\footnote{Reference code is available at: \url{https://github.com/XplaiNLP/SpecFi-Narrative-Retrieval}} operates in two variants: SpecFi-DR retrieves the nearest text from the reference corpus via dense retrieval as a few-shot example. SpecFi-CS retrieves high-level community summaries via graph-based search over a heterogeneous knowledge graph \citep{2025noderag}. Our evaluation shows that the community summaries improve performance beyond what actual samples from the training set achieve. Our analysis further reveals that these summaries can converge on descriptions close to expert-crafted narrative taxonomies like CARDS \citeplanguageresource{lr-CARDS}, suggesting that graph-based methods can extract narrative structure from unlabeled text; a property with application for monitoring emerging narratives.

\item \textbf{Narrative variance as a predictor of retrieval difficulty.} We propose narrative variance ($V_i$), an embedding-based metric quantifying the internal spread of texts within a narrative group, and show via partial correlation analysis (controlling for sample size) that it correlates with retrieval difficulty for standard sparse and dense baselines. SpecFi-CS shows the smallest sensitivity to this effect, maintaining stable performance across high-variance narratives.

\end{enumerate}

\section{Preliminaries and Related Work}
\subsection{Disinformation Narrative Classification and Retrieval}

Several recently released works organize disinformation texts under narrative taxonomies on different topics \citep{kotseva_trend_2023, sosnowski-etal-2024-eu, 2025ukelect, heinrich-etal-2024-automatic}. Our focus in this paper is specific to climate change denial narrative datasets (CARDS \citeplanguageresource{lr-CARDS}), climate obstruction in social media advertising (CO, \citeplanguageresource{lr-CO}), and climate disinformation in news (PolyNarrative, \citeplanguageresource{lr-PN}, which consists of two topic splits: Climate Change and War in Ukraine).

Within related domains, the term ``narrative retrieval'' is used mainly to describe claim retrieval in practice, focusing on individual claims, not overarching elements like core messages, such as in \citet{singh2024breakinglanguagebarriersmmtweets, singh2024debunked}. \citet{akter2024fansfacetbasednarrativesimilarity} identified the need for metrics that capture narrative similarity beyond surface representations, and \citet{hatzel-biemann-2024-story} demonstrated the difficulty of narrative retrieval by showing that untailored dense retrieval substantially underperforms on the task of retrieving texts by their summaries.

\begin{figure*}[t]
    \centering
    \includegraphics[scale=0.66]{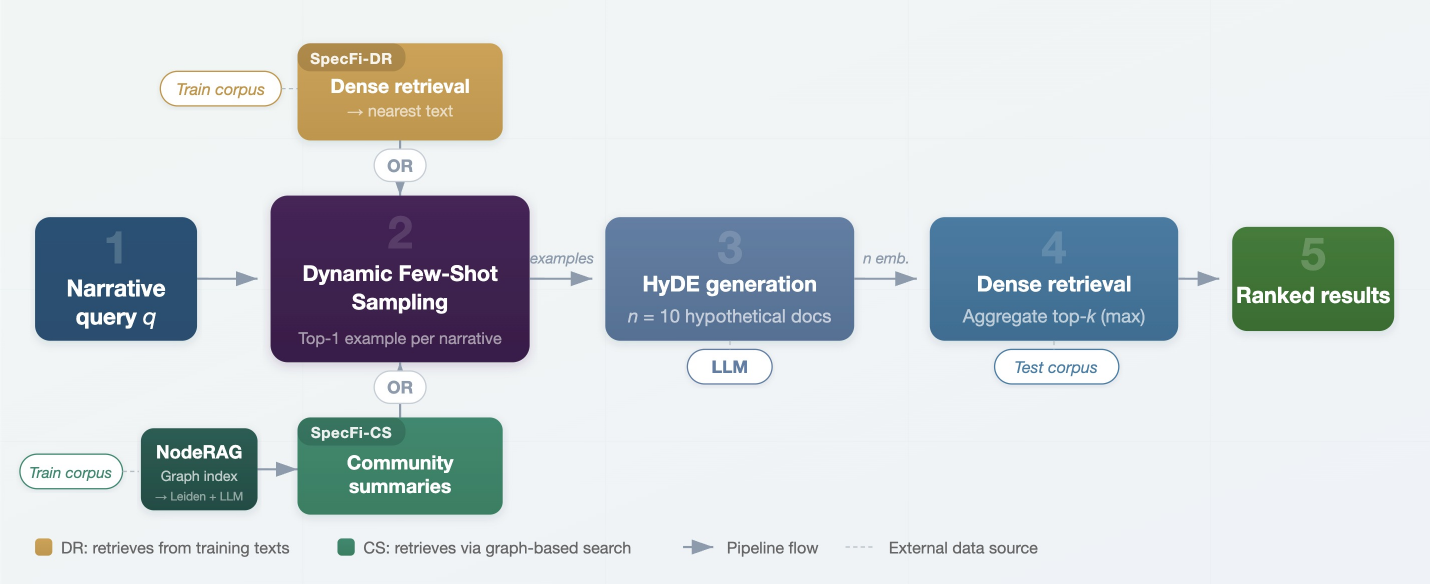}
    \caption{Overview: The SpecFi Retrieval Pipeline}
    \label{fig:pipeline}
\end{figure*}

\paragraph{Hypothetical Document Embeddings}

The retrieval strategy of generating hypothetical documents to bridge the gap between query and document representations was introduced as HyDE by \citet{2023hyde}. Given a query, HyDE generates $n$ hypothetical documents, embeds them, and retrieves based on the aggregated embeddings. This effectively expands queries with vocabulary and semantic meaning which is representative for relevant documents, which is a valuable property for narrative retrieval, where narrative descriptions are abstract while their instantiations are concrete. This generation step can be understood as a computational analogue of what \citet{Roine2020} calls the instrumental mode of speculation: generating possible consequences from a given premise. We adopt this framing in our system name (SpecFi, Speculative Fiction).

\paragraph{Graph-Based Reasoning in Retrieval}

Narratives can often be understood as sets of narrative features and their interrelated structures \cite{2021piper, Hellman2024}. Since embedding-based similarity can fail at capturing complex relational structures (as discussed in the introduction) graph-based representations offer an alternative: they can explicitly model entities, relationships, and thematic co-occurrence patterns. Within current retrieval research, several recent graph-based RAG frameworks construct knowledge graphs from unstructured corpora and apply community detection to identify thematic clusters. GraphRAG \citep{2025graphrag} introduced this paradigm: an LLM extracts entities and relationships, the Leiden algorithm \citep{Traag2019} partitions the resulting graph into hierarchically nested communities, and a second LLM pass generates bottom-up summaries for each community. These summaries serve as coarse semantic layers for query-focused summarization at retrieval time. 
We build on NodeRAG \citep{2025noderag}, which refines this approach by operating over a heterogeneous graph with a search pipeline that propagates relevance through the graph structure; details are given in section \ref{sec:pipeline}.

\section{Methodology}

\subsection{Retrieval Pipeline}
\label{sec:pipeline}
The retrieval pipeline, illustrated in Figure~\ref{fig:pipeline}, operates in five steps. We model a narrative monitoring scenario in which an analyst queries a text corpus by core message to identify texts aligned with a given narrative. For each dataset, we use the training split as a reference corpus and the test split as the evaluation set; narrative descriptions derived from each dataset's taxonomy serve as query proxies (collected in our code repository). The reference corpus is used without access to narrative labels, simulating a realistic setting in which previously collected texts (including non-disinformation content) are available but lack narrative annotations.

\begin{algorithm}[h!]
\caption{SpecFi Narrative Retrieval Workflow}
\label{alg:specfi}
\begin{algorithmic}[1]
\Require Narrative taxonomy with labels used as queries $\{q_1, \dots, q_K\}$, reference corpus $t$, NodeRAG graph index $\mathcal{G}$
\Ensure Ranked list of candidate texts per narrative
\State Select target narrative $q_k$
\State For each $q_k$, retrieve one example via:
    \begin{enumerate}[label=(\alph*)]
        \item SpecFi-DR: nearest text from $t$ by cosine similarity, or
        \item SpecFi-CS: top-ranked high-level element from $\mathcal{G}$ via NodeRAG's graph-based search
    \end{enumerate}
 and concatenate all $K$ narrative-example pairs as few-shot examples
\State For target $q_k$, generate $n{=}10$ hypothetical documents
\State Embed hypotheticals; retrieve top-$k$ from test set via aggregated dense retrieval
\State Return ranked results
\end{algorithmic}
\end{algorithm}

NodeRAG \citep{2025noderag} constructs a heterogeneous graph from the input corpus comprising seven node types, including entities, relationships, semantic units, and text chunks. During the graph augmentation stage, the Leiden community detection algorithm \citep{Traag2019} is applied to segment the graph into communities of closely related nodes. For each detected community, an LLM analyzes the aggregated content of its member nodes and generates high-level element nodes which are represented by community summaries. These high-level elements are reintegrated into the graph, providing a summarization layer that captures patterns beyond what is present in any individual text. At query time, NodeRAG's search combines embedding similarity and entity matching to identify seed nodes, then propagates relevance scores through the heterogeneous graph via Personalized PageRank. This means that a high-level element can be surfaced not only through direct similarity to the query but also through structural connectivity to other relevant nodes.
In our SpecFi-CS pipeline, we query this search pipeline with each narrative description and extract the top-ranked high-level element from the retrieval results, using it as a few-shot example for hypothetical document generation.
This exploits the summaries' abstracted nature to produce hypotheticals that span the interpretive range of a narrative rather than anchoring on a single text. For each narrative, we generate $n{=}10$ hypothetical documents, selected based on preliminary experiments.

\paragraph{Illustrative Example}
Consider the CARDS narrative \emph{``Climate impacts / global
warming is beneficial / not bad. CO$_2$ is beneficial / not a
pollutant. CO$_2$ is plant food''} (narrative id: 3\_3).
\\
\noindent\textbf{SpecFi-DR} retrieves the nearest text from the reference corpus as a few-shot example:
\begin{quote}\small
``Idso pointed out that there is a huge body of literature on the biological impacts of rising temperatures and atmospheric CO2 levels 
that the International Panel on Climate Change (IPCC) ignores. He emphatically stated that atmospheric CO2 is not a pollutant. In fact, 
increased levels of CO2 reduce the negative effects of a number of plant stresses [\ldots] and protects against herbivores.''
\end{quote}

\noindent\textbf{SpecFi-CS} instead retrieves community summary:
\begin{quote}\small
``Some argue that the effects of CO2 increases and slight global warming may be harmless or even beneficial, challenging alarmist narratives about climate change.''
\end{quote}

Notably, the community summaries are generated without access to narrative labels; we discuss their convergence with the expert-crafted taxonomy in section \ref{sec:discussion_cs}.

\subsection{Datasets}

\begin{table*}[t] 
\centering
\resizebox{1\textwidth}{!}{
\begin{tabular}{l|ccccc|cccc}
            \toprule

  &  narratives &  mean texts  &  std texts  & mean words  &  std words  &  mean words  &  std words &  total  &  disinfo \% \\
 & $n$  &  per $n$  & per $n$  &  per $n$ &  per $n$ &  per text  &  per text  &  texts  &  of all texts \\
\midrule
CARDS  & 17 & 67.65 & 57.49 & 7.61 & 3.96 & 65.35 & 57.60  & 2904 & 39.6 \\
CO & 7 & 38.29 & 27.20 & 20.50 & 3.56 & 28.27 & 11.95  & 255 & 73.3 \\
PN & 51 & 1.98 & 1.39 & 9.06 & 2.82 & 601.83 & 293.73  & 41 & 73.2 \\
\midrule
PN-UKR & 27 & 2 & 1.47 & 9.32 & 3.30 & 740.62 & 382.17  & 13 & 100 \\
PN-CC  & 23 & 2 & 1.32 & 8.77 & 2.12 & 495.71 & 120.74  & 17 & 100 \\
PN-Neutral  & 0 &  -  &  -  &  -  &  -  & 459.78 & 166.97  & 11 & 0 \\
            \bottomrule
        \end{tabular}
            }
\caption{Quantitative statistics of the used datasets. 
}
\label{tab:datasets_stats}
        \end{table*}

\paragraph{CARDS}
The Computer-Assisted Recognition of (Climate Change) Denial and Skepticism dataset \citeplanguageresource{lr-CARDS} contains climate change denial claims organized under a two-level taxonomy of 5 main narratives and 27 subnarratives, of which 17 are attested in the data. Each text is a short claim (mean 65 words) mapped to one narrative. With 2,904 texts in the test set and 21-225 texts per narrative, CARDS provides the densest evaluation setting and is the primary dataset for our statistical analysis.

\paragraph{Climate Obstruction (CO)}
The Climate Obstruction dataset \citeplanguageresource{lr-CO} contains social media advertisements from oil and gas companies, classified under 7 obstruction narratives such as corporate community engagement and clean energy leadership. Here, the texts are shorter (mean 28 words), may carry multiple labels and are designed to reshape public perception of the fossil fuel industry.

\paragraph{PolyNarrative Climate Change Subset (PN-CC)}
The PolyNarrative dataset \citeplanguageresource{lr-PN} contains news articles annotated with fine-grained narrative
labels across multiple topics. For better comparability, we use the English language climate change related subset. Texts are substantially longer (mean 496 words) and frequently carry multiple narrative labels.
With only 56 climate-related texts in the development set (used as test set; labels were not released for the actual test split), PN-CC serves as a complementary low-resource evaluation but does not support reliable statistical analysis.

The three datasets differ across several dimensions relevant to narrative retrieval evaluation, allowing us to test whether SpecFi generalizes across the heterogeneous landscape of climate disinformation. Table \ref{tab:datasets_stats} summarizes quantitative statistics of the datasets.

Narrative descriptions used as queries are constructed from each dataset's taxonomy by concatenating hierarchical labels (e.g., for CARDS: ``Global warming is not happening. Ice/permafrost/snow cover isn't melting''). 

\subsection{Metrics}
\label{sec:metrics}

\paragraph{Retrieval Performance}

We report Mean Average Precision (MAP), which summarizes precision across all recall levels; normalized Discounted Cumulative Gain at cutoffs 10 and 100 (nDCG@10, nDCG@100), which measures ranking quality with position-based discounting; and average R-Precision, the precision at the rank equal to the number of relevant documents. All are standard information retrieval metrics \citep{manning2008introduction}. Each evaluation is performed over $K$ narratives per dataset, yielding $K$ per-narrative scores that we aggregate by macro-averaging.

\paragraph{Embedding-Based Narrative Metrics}

Let $\mathcal{N} = \{n_1, \dots, n_K\}$ be a set of narratives.
Each narrative $n_i$ has an associated set of text embeddings 
$\mathcal{T}_i = \{\mathbf{t}_{i1}, \dots, \mathbf{t}_{im_i}\} \subset \mathbb{R}^d$
with centroid $\mathbf{c}_i = \frac{1}{m_i}\sum_{j=1}^{m_i} \mathbf{t}_{ij}$.
We define cosine distance as 
$d_{\cos}(\mathbf{a}, \mathbf{b}) = 1 - \frac{\mathbf{a} \cdot \mathbf{b}}
{\lVert\mathbf{a}\rVert\,\lVert\mathbf{b}\rVert}$.

\medskip\noindent
\textbf{Narrative Distinctness}, as proposed in \citet{2025narrdistinct}, measures how separable a narrative is from the others via inter-centroid distances $d_{ij} = d_{\cos}(\mathbf{c}_i, \mathbf{c}_j)$. The geometric mean balances global separation (mean distance) with local distinctiveness (minimum distance):

\begin{align}
  D_i &= \sqrt{\bar{d}_i \cdot d_i^{\min}}, \label{eq:distinct}\\
  \bar{d}_i &= \tfrac{1}{K{-}1}\textstyle\sum_{j \neq i} d_{ij}, \quad
  d_i^{\min} = \min_{j \neq i}\, d_{ij}. \nonumber
\end{align}


\noindent
\textbf{Narrative Variance} measures the overall spread of texts 
around the centroid:
\begin{equation}
  V_i = \frac{1}{m_i}\sum_{j=1}^{m_i} 
  \lVert \mathbf{t}_{ij} - \mathbf{c}_i \rVert_2^2.
\end{equation}

\noindent
The two metrics operationalize different aspects of the notion of measuring a narrative's interpretation space: $D_i$ captures how separable this space is from neighboring narratives and $V_i$ captures the overall spread of instantiations around the narrative's center. We treat them as competing hypotheses about what drives retrieval difficulty: is it proximity to other narratives ($D_i$) or overall internal diversity ($V_i$)? We test this in section \ref{sec:stats}.

\subsection{Model Choice}
\label{sec:models}

For hypothetical document generation, we use \texttt{gpt-4o}
\citep{openai2024gpt4ocard} and \texttt{gemma-3-27b-it}
\citep{gemmateam2025gemma3technicalreport} (including an uncensored or ``abliterated'' variant with safety alignment removed in post-training to mitigate possible refusals when generating disinformation texts, denoted~\texttt{-a}). The models are run as Q8\_0 GGUF quantizations. For dense retrieval embeddings, we use Qwen3-Embedding-4B \cite{yang2025qwen3technicalreport} due to its strong performance on MTEB\footnote{\url{https://huggingface.co/spaces/mteb/leaderboard}} and support for instruction prompts. For the embedding-based narrative metrics ($D_i$, $V_i$), we use \textsc{gte-large} \citep{li2023gte} ($d = 1024$) to separate the analysis from the retrieval pipeline. NodeRAG graph construction follows the framework's default configuration with OpenAI models for structured output generation.

\section{Evaluation}

\begin{table}[!t]
    \centering
    \resizebox{0.48\textwidth}{!}{
        \begin{tabular}{l|l|ccccc}
            \toprule
 &   &  & NDCG & NDCG &  Average \\
Setup  &  Models & MAP & @10 & @100 &  R-Precision \\
\midrule
zero  & 4o, OI-E & 0.321 & 0.509 & 0.487 &  0.370  \\
 shot & G,Q4B & 0.313 & 0.469 & 0.456  & 0.371  \\
 & G-a,Q4B & 0.295 & 0.428 & 0.436 &  0.308  \\
\midrule
static* &  4o, OI-E  & 0.488 & 0.713 & 0.649 & 0.487  \\
 & G, Q4B & 0.435 & 0.635 & 0.616 &  0.435  \\
 & G-a, Q4B & 0.464 & 0.679 & 0.637 &  0.468  \\
\midrule
SpecFi   & 4o, OI-E & 0.421 & 0.682 & 0.600 &  0.440  \\
 -DR& G, Q4B & 0.424 & 0.630 & 0.581 &  0.432  \\
  &  G-a, Q4B & 0.457 & 0.693 & 0.619 & 0.453  \\
\midrule
SpecFi   &  4o, OI-E & 0.426 & 0.660 & 0.597 & 0.456 \\
 -CS & G, Q4B & 0.468 & 0.709 & 0.631 & \textbf{0.492} \\
  & G-a, Q4B & \textbf{ 0.494} &  \textbf{0.726 } & \textbf{ 0.657} & 0.487   \\
            \bottomrule
        \end{tabular}
    }
\caption{Results on the CARDS dataset. static* is included for reference only due to its reliance on labels. All metrics are averaged over 10 runs. We report a standard deviation of <0.01 for all performance metrics. The model abbreviations are: 4o=gpt-4o, OI-E= text-embedding-3-large, G=gemma-3-27b-it, G-a=gemma-3-27b-it abliterated, Q4B=Qwen3-Embedding-4b.}  
\label{tab:main}
\end{table}

\paragraph{Retrieval Performance}
We first evaluate our system on performance metrics to further analyze possible correlations with the narrative metrics introduced above. Table \ref{tab:main} documents our results, where averages of metrics over 10 runs are presented due to randomized factors within HyDE. On CARDS, SpecFi-CS with the abliterated model achieves the highest MAP (0.494) among all label-free setups, outperforming both the dense baseline (0.299) and SpecFi-DR (0.457). On CO, SpecFi-DR outperforms SpecFi-CS (0.519 vs.\ 0.491), suggesting that the relative advantage of community summaries over retrieved texts depends on dataset characteristics. For comparison, we also include the setups labeled with ``static'' where few-shot examples were statically retrieved by assessing the labels.


\begin{table}[h]
    \centering
    \resizebox{0.4\textwidth}{!}{ 
        \begin{tabular}{llccccc}
             \toprule
 &   &     NDCG & Avg.   \\
Setup/Model  & MAP & @10 &  R-Prec. \\
\midrule
BM25 & 0.326 & 0.472 & 0.298  \\
Qwen3-E-4B & 0.499 & 0.607 & 0.491  \\
\midrule
SpecFi-DR & \textbf{0.519} & \textbf{0.644} & \textbf{0.496}  \\
SpecFi-DR-a & 0.482 & 0.604 & 0.494\\
SpecFi-CS & 0.491 & 0.618 & 0.49  \\
SpecFi-CS-a & 0.495 & 0.627 & 0.486  \\
             \bottomrule
        \end{tabular}
    }
    \caption{Evaluation on CO. Qwen3-E-4B=Qwen3-Embedding-4B}
    \label{tab:co}
\end{table}

\begin{table}[h]
    \centering
    \resizebox{0.4\textwidth}{!}{ 
        \begin{tabular}{lcccccc}
             \toprule
 &   &     NDCG & Avg.  \\
Setup/Model  & MAP & @10 &  R-Prec.  \\
\midrule
BM25 & 0.311 & 0.378 & 0.219  \\
Qwen3-E-4B & \textbf{0.502} & 0.598 & 0.374  \\
\midrule

SpecFi-DR & 0.443 & 0.621 & 0.370 \\
SpecFi-DR-a  & 0.386 & 0.536 & 0.275 \\
SpecFi-CS & 0.458 & 0.626 & 0.372  \\
SpecFi-CS-a & 0.471 & \textbf{ 0.640} & \textbf{0.386} \\

            \bottomrule
        \end{tabular}
    }
    \caption{Evaluation on PN}
    \label{tab:pn}
\end{table}




\paragraph{Component Analysis}

To further analyze the influence of the components of our system, we run different ablation studies documented in Table \ref{tab:ablation}. Here, our main goal is to provide comparison between the proposed SpecFi setups and results from sparse and dense retrieval only, since these performance metrics are also the base for our statistical analysis of correlation. We also include results for NodeRAG only, where we patched the framework to retrieve the full list of top $k$ results directly.
To isolate the contribution of hypothetical document generation, we evaluate CS-direct, which uses the community summary as a direct query expansion without any generation step. CS-direct achieves a MAP of 0.357, above the dense baseline (0.299) but substantially below SpecFi-CS-a (0.494), indicating that the community summaries provide modest retrieval benefit as query expansions but that the majority of SpecFi-CS's performance gain is attributable to the speculative generation step.

\begin{table}[t]
\centering
\resizebox{0.5\textwidth}{!}{
\begin{tabular}{lccccc}
\toprule
& & NDCG & Avg.  & s / \\
Setup/Model & MAP & @10 & R-Prec.  & narrative \\
\midrule
NodeRAG only & 0.259 & 0.506 & 0.323  & 1.931 \\
\midrule
BM25 & 0.080 & 0.125 & 0.119 &  0.011 \\
thenlper/gte-large & 0.215 & 0.394 & 0.272 &  2.092 \\
OpenAI-E & 0.262 & 0.507 & 0.323 &  0.452 \\
\midrule
Qwen3-E-4B & 0.299 & 0.523 & 0.352 & 6.645 \\
Qwen3-E-4B-p & 0.316 & 0.536 & 0.370 &  6.593 \\
\midrule
CS-direct & 0.357 & 0.536 & 0.370 & 1.300 \\
\midrule
SpecFi-CS-a & 0.494 &  0.726& 0.487 & 14.80 \\
\bottomrule
\end{tabular}
}
\caption{Retrieval performance of individual pipeline components on CARDS, serving as baselines for the statistical analysis in §5. Models: OpenAI-E=text-embedding-3-large, Qwen3-E-4B=Qwen3-Embedding-4b. Runtimes were measured on a system with a H100 GPU.}
\label{tab:ablation}
\end{table}


\paragraph{Refusal and Abliteration Analysis.}
To assess whether the Gemma models refused to generate disinformation-aligned texts, we scanned all generated hypothetical documents ($n{=}170$ per model) for refusal indicators including direct refusals, role-breaking statements, and safety-related language. Neither the abliterated (G-a) nor the non-abliterated (G) variant produced any refusals (0\% refusal rate). However, the two models differ in output length: G produces longer texts in 110 out of 170 paired generations (mean 48.0 vs.\ 41.2 words). Since HyDE retrieval relies on cosine similarity between generated and corpus texts in embedding space, we hypothesize that the abliterated model's more concise outputs favor direct claims over verbose qualifications and yield embeddings closer to the shorter, assertive texts typical of disinformation samples in CARDS, consistent with the performance advantage of G-a over G observed across all few-shot configurations in Table~\ref{tab:main}.

\paragraph{Number of Hypothetical Documents}
We ablated $n \in \{1, 5, 10, 20\}$ for SpecFi-CS-a on CARDS to evaluate the influence on retrieval performance. MAP increases from 0.438 ($n{=}1$) to 0.484 ($n{=}5$) and plateaus at 0.494 ($n{=}10$) and 0.491 ($n{=}20$), while runtime scales approximately linearly in $n$, making $n{=}10$ a practical tradeoff between retrieval performance and computational cost.

\paragraph{Exploratory Transfer to CO and PN-CC.}

We further compare performance metrics (Table \ref{tab:co} and Table \ref{tab:pn}) and possible correlations (Table \ref{tab:stats}) on two other datasets. 

\begin{table}[t]
    \centering
    \resizebox{0.4\textwidth}{!}{
        \begin{tabular}{llcc}
            \toprule
 & & Narrative & Narrative \\
Setup & Dataset & Distinct. & Variance \\
\midrule
\multirow{3}{*}{BM25}
& CARDS & -0.240 & -0.525\textsuperscript{*} \\
& CO & -0.357 & -0.071 \\
& PN & 0.369\textsuperscript{*} & 0.319\textsuperscript{*} \\
\midrule
\multirow{3}{*}{QWEN-E-4B}
& CARDS & -0.066 & -0.556\textsuperscript{*} \\
& CO & -0.679 & 0.000 \\
& PN & 0.197 & 0.151 \\
\midrule
\multirow{3}{*}{SpecFi-DR-a}
& CARDS & 0.147 & -0.578\textsuperscript{*} \\
& CO & -0.964 & 0.214 \\
& PN & -0.016 & 0.476\textsuperscript{**} \\
\midrule
\multirow{3}{*}{SpecFi-CS-a}
& CARDS & 0.282 & -0.324 \\
& CO & -0.786\textsuperscript{**} & -0.286 \\
& PN & -0.041 & 0.249 \\
\bottomrule
        \end{tabular}
    }
    \caption{Spearman's $\rho$ between MAP and narrative metrics. FDR-corrected significance: \textsuperscript{*}p<0.05, \textsuperscript{**}p<0.01.}
    \label{tab:stats}
\end{table}

\section{Statistical Analysis}
\label{sec:stats}

For all tests, we compute Spearman's $\rho$ with FDR correction following the Benjamini--Hochberg procedure.
Table~\ref{tab:stats} reports correlations between MAP and both narrative metrics across datasets. We treat these metrics as competing operationalizations of a narrative's interpretive space and ask which, if any, is associated with retrieval difficulty.

On CARDS, narrative variance shows consistent negative correlations with MAP across all four systems, reaching significance for BM25, QWEN-E-4B, and SpecFi-DR-a (Table~\ref{tab:stats}). Narrative distinctness does not reach significance on CARDS in the uncontrolled analysis, suggesting that retrieval difficulty is driven by the overall embedding spread within a narrative rather than by inter-narrative separation ($D_i$). On CO, correlations should be interpreted with caution given the limited number of narratives ($K = 7$); the only significant result is a negative correlation between narrative distinctness and SpecFi-CS-a ($\rho = -0.786$, $p < 0.01$). On PN, the positive correlations between MAP and narrative variance (e.g., BM25: $\rho = +0.319$; SpecFi-DR-a: $\rho = +0.476$) run opposite to the pattern observed on CARDS. We attribute this reversal to two properties of the PN dataset: per-narrative sample sizes are very small (mean $m_i = 2$), making variance estimates unreliable, and the multi-label annotation structure conflates intra-narrative spread with cross-narrative overlap. We therefore restrict our narrative metric analysis to CARDS, where per-narrative sample sizes ($m_i \in [21, 225]$) support reliable estimation. Leave-one-out analysis confirms that no single narrative, including those with the smallest sample sizes, drives the observed correlations on CARDS.

\begin{figure*}[h]
    \centering
    \includegraphics[scale=0.65]{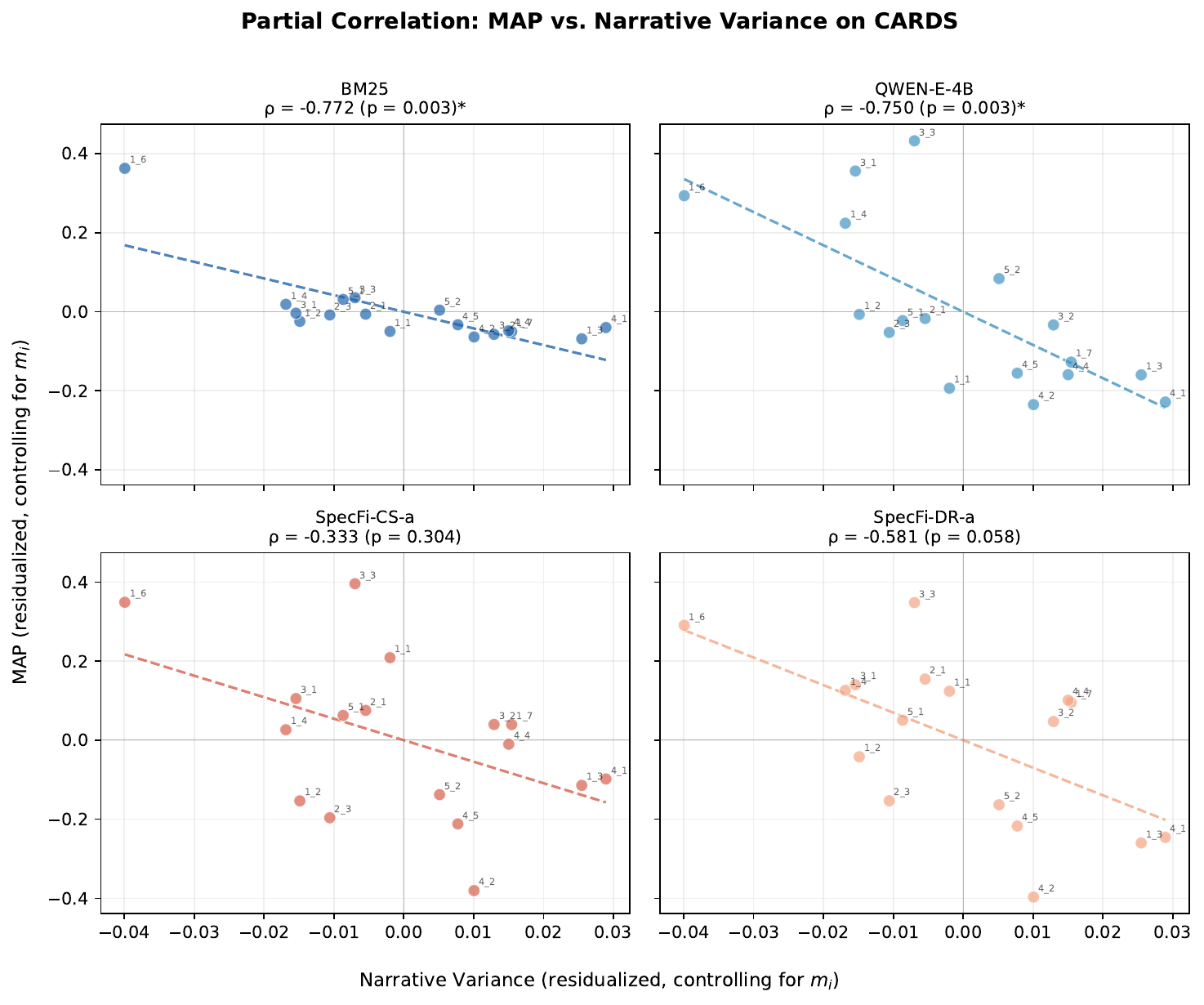}
    \caption{Partial correlation between MAP and narrative variance on CARDS, controlling for $m_i$. Each point
represents one narrative. BM25 and QWEN-E-4B show steep negative slopes; SpecFi-CS-a shows no significant trend. All $p$-values are FDR-corrected (Benjamini--Hochberg across all tests in
Table~\ref{tab:partial_corr}).}
    \label{fig:scatter_map_vs_variance}
\end{figure*}

\begin{figure}[h]
    \centering
    \includegraphics[scale=0.385]{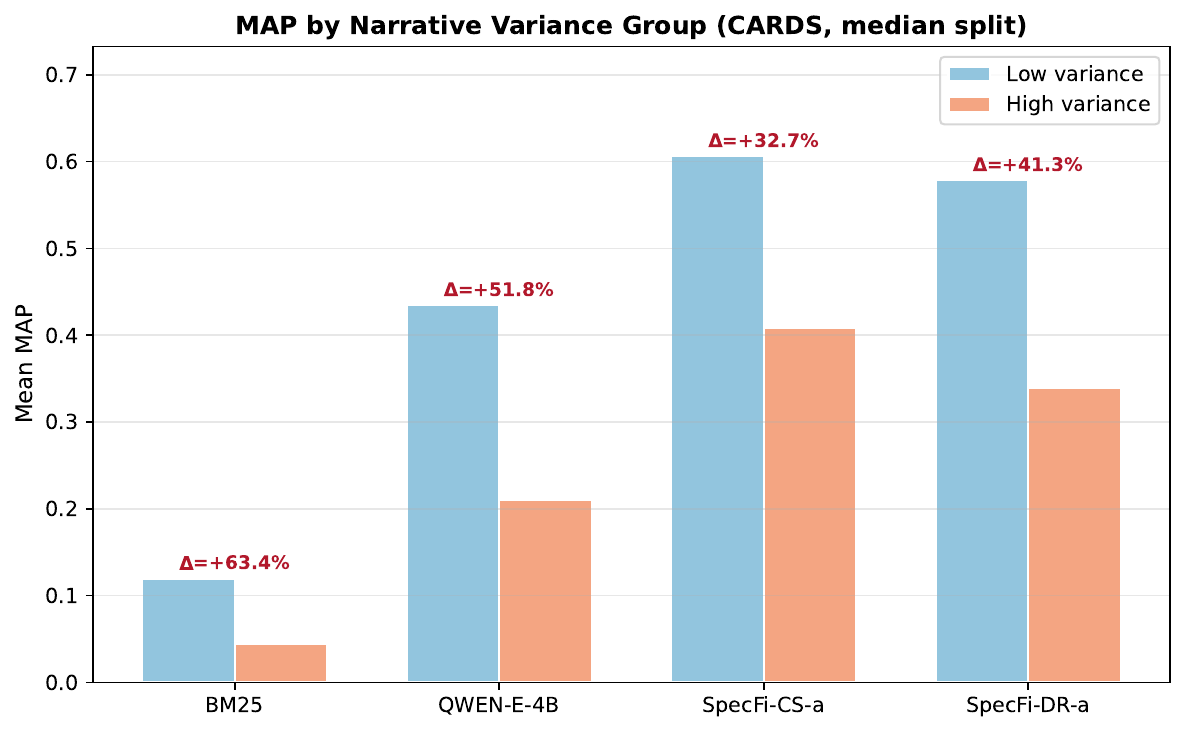}
    \caption{Mean MAP for narratives above and below median $V_i$ on CARDS. BM25 loses 63.4\% of its MAP on high-variance
    narratives; SpecFi-CS-a loses 32.7\%.}
    \label{fig:stratified_map}
\end{figure}


\begin{table}[H]
\centering
\resizebox{0.4\textwidth}{!}{
\begin{tabular}{lcc}
\toprule
\multicolumn{3}{l}{\textit{Original correlations}} \\
\midrule
Setup & $D_i$ & $V_i$ \\
\midrule
BM25        & -0.240 (.530) & -0.525 (.123) \\
QWEN-E-4B   & -0.066 (.874) & -0.556 (.122) \\
SpecFi-DR-a  & ~~0.147 (.704) & -0.578 (.122) \\
SpecFi-CS-a  & ~~0.282 (.468) & -0.324 (.468) \\
\midrule
\multicolumn{3}{l}{\textit{Partial correlations (controlling for $m_i$)}} \\
\midrule
Setup & $D_i$ & $V_i$ \\
\midrule
BM25        & -0.029 (.978) & -0.772 (\textbf{.003}) \\
QWEN-E-4B   & -0.007 (.978) & -0.750 (\textbf{.003}) \\
SpecFi-DR-a  & ~~0.125 (.759) & -0.581 (.058) \\
SpecFi-CS-a  & ~~0.387 (.249) & -0.333 (.304) \\
\bottomrule
\end{tabular}
}
\caption{Spearman's $\rho$ between MAP and narrative metrics on CARDS. FDR-corrected $p$-values; \textbf{bold} $p < 0.05$.}
\label{tab:partial_corr}
\end{table}

\paragraph{Controlling for Sample Size}
The number of texts per narrative ($m_i$) varies from 21 to 225 on CARDS and may itself correlate with both MAP and narrative metrics. We compute partial Spearman correlations by residualizing both MAP and each metric against $m_i$ via linear regression. Table~\ref{tab:partial_corr} reports results for both metrics; Figure~\ref{fig:scatter_map_vs_variance} visualizes the relationship for narrative variance. 
Here, the partial correlations strengthen relative to the uncontrolled analysis: BM25 moves from $\rho = -0.525$ to $\rho = -0.772$ and QWEN-E-4B from $\rho = -0.556$ to $\rho = -0.750$, both significant
after FDR correction ($p = 0.003$). SpecFi-DR-a shows a borderline effect ($\rho = -0.581$, $p_{\text{FDR}} = 0.058$; raw $p = 0.014$), significant in all 17 LOO iterations but not after FDR correction; while SpecFi-CS-a remains non-significant ($\rho = -0.333$, $p_{\text{FDR}} = 0.304$). Two-tailed permutation tests ($10{,}000$ iterations) confirm these results ($p_{\text{perm}} = 0.0007$, $0.0011$, $0.014$, and $0.196$ for BM25, QWEN-E-4B, SpecFi-DR-a, and SpecFi-CS-a, respectively).

Narrative distinctness remains non-significant throughout. Together, these results indicate that between the two embedding-based narrative metrics, it is the overall intra-narrative spread ($V_i$), not inter-narrative separation ($D_i$), that correlates with retrieval difficulty. This is consistent with the interpretation that standard retrieval degrades when a narrative manifests through many diverse framings, rather than when it is merely close to neighboring narratives in embedding space. A median split on $V_i$ (Figure~\ref{fig:stratified_map}) quantifies this effect: BM25 loses 63.4\% of its MAP when moving from low- to high-variance narratives, QWEN-E-4B loses 51.8\%, and SpecFi-DR-a loses 41.3\%. SpecFi-CS-a shows the smallest degradation (32.7\%) while maintaining the highest absolute MAP in both groups. Leave-one-out analysis confirms stability: partial correlations remain significant in all 17 iterations for BM25, QWEN-E-4B, and SpecFi-DR-a, with no single narrative acting as a leverage point (BM25 LOO range: $\rho \in [-0.83, -0.73]$). 

\section{Discussion}
\label{sec:discussion_cs}

Analysis of the community summaries retrieved for each CARDS narrative reveals a key insight: the community summaries are generated without access to narrative labels, NodeRAG constructs its knowledge graph and applies Leiden community detection exclusively on the textual content of the training corpus. Of the 17 CARDS narratives, 11 receive summaries that align with the taxonomy label at least at the super-claim level, 2 are collapsed with a sibling sub-narrative, and 4 exhibit drift or incoherence (full mapping is provided in Table \ref{tab:summaries} in the appendix). For instance, the summary retrieved for narrative 1\_2 (``heading into ice age / global cooling'') independently arrives at ``the Earth may be entering a cooling cycle,'' and narrative 5\_1 (``science is uncertain / unreliable'') yields ``skepticism about the reliability of climate models.'' This convergence suggests that the CARDS narrative taxonomy reflects genuine topical structure in the disinformation corpus rather than predefined classification, and that graph-based community detection can surface this structure from unlabeled text; a property with application for monitoring emerging narratives that lack predefined labels.

Instances where the summaries fail reveal diagnostic patterns that help explain the system's behavior. The collapse pattern (where sub-narratives such as 4\_1/4\_2 or 3\_2/3\_3 receive identical summaries) correlates directly with low per-narrative AP for SpecFi-CS and identifies Leiden resolution as a tunable parameter. The drift pattern reveals a subtler problem: narrative 4\_4 (``clean energy won't work'') receives a summary that argues \emph{for} technological solutions, inverting the narrative's stance. This polarity inversion arises because community detection clusters texts by topic co-occurrence, which does not inherently distinguish argumentative direction. Texts criticizing and texts promoting renewable energy share entities and relationships (solar panels, wind turbines, efficiency, cost), so Leiden groups them together, and the LLM's summary reflects the majority framing. This failure mode suggests that strategies like stance-aware community detection could address a class of errors that finer resolution alone would not resolve.

These failure patterns are consistent with SpecFi-CS-a showing the lowest inter-system correlation with BM25 on CARDS ($\rho = 0.365$, $p = 0.249$), indicating that the community-summary-based system retrieves narratives through a qualitatively different mechanism than dense or lexical retrieval, producing complementary errors. Where community summaries converge on the correct narrative premise, SpecFi-CS generates hypotheticals that span the narrative's interpretive range---as illustrated by narrative 3\_3 (``CO$_2$ is plant food''), where the abstract summary enables generation of diverse hypothetical documents covering CO$_2$ fertilization, agricultural productivity, and pollutant classification arguments, rather than anchoring on a single text's framing. Where summaries collapse or drift, the generated hypotheticals lose discriminative power or target the wrong stance entirely. 

\section{Future Work}

While this study focuses on climate change denial, the SpecFi framework is domain-agnostic. Applying it to other narrative datasets (such as European disinformation narratives \citep{sosnowski-etal-2024-eu}, COVID-19 conspiracy narratives \citep{heinrich-etal-2024-automatic}, or propaganda taxonomies \citep{Solopova2023b, sahitaj-etal-2025-hybrid}) would test the generalizability of both the retrieval approach and the narrative variance metric. On the retrieval side, the final step still relies on dense cosine similarity. Following \citet{hatzel-biemann-2024-story} and \citet{akter2024fansfacetbasednarrativesimilarity}, more interpretive similarity measures that incorporate narrative features such as actors, localities, and argumentative structure could be explored. Similarly, aligning the graph representation more closely with narrative systems \citep{Hellman2024} could improve both community summary quality and retrieval performance.

\section{Conclusion}

In this study, we re-framed climate disinformation detection as a narrative retrieval task and introduced SpecFi, a speculative-document generation framework that bridges abstract narrative descriptions and their diverse textual realizations. Across three datasets, SpecFi, and especially the community-summary variant, improves robustness compared to sparse and dense baselines, remaining stable even for high-variance narratives. Our analysis further shows that narrative variance correlates with retrieval difficulty for standard baselines, while graph-derived community summaries can recover narrative structure from unlabeled data. Together, these results highlight narrative retrieval as a flexible approach for tracking evolving disinformation narratives beyond fixed taxonomies.

\section*{Limitations}

While we were able to provide a version of SpecFi-DR which only relies on open source models to ensure reproducibility, the SpecFi-CS setups include one reliance on OpenAI models within NodeRAG. Recent studies have shown that OpenAI models still outperform on structured output generation \cite{geng2025jsonsch}, which is an essential step during graph construction. For this reason and due to NodeRAG's own recommendation\footnote{\url{https://terry-xu-666.github.io/NodeRAG_web/blog/2025/03/16/structure-output/}}, we used the proprietary model here.
An additional factor that could affect our results: the CARDS dataset is from 2021, making it likely to be part of the training data of LLMs. While this does not necessarily relate to our specific usage of this dataset, it is still possible that there is an influence on the generation of hypothetical documents as well as community summaries. However, our results of the zero shot variants in Table \ref{tab:main} indicate that none of our tested LLMs is capable of generating representative hypotheticals without examples and only based on the narrative by itself, but an influence in some kind of capacity cannot be ruled out. 
Our evaluation relies on automatic retrieval metrics derived from existing narrative annotations; human evaluation of narrative alignment quality remains for future work. Similarly, the convergence analysis between community summaries and expert-crafted taxonomies (Section~6) is based on qualitative judgment. We provide a systematic mapping of all 17 narratives to pattern categories in Table~\ref{tab:summaries} in the appendix for verification, but a more rigorous evaluation with independent annotators would strengthen this claim.

\section*{Ethical Considerations}
Recent work has shown that current LLMs can generate convincing disinformation
following predefined narratives \cite{vykopal-etal-2024-disinformation} and that personalization
requests can bypass safety filters \cite{zugecova-etal-2025-evaluation}, highlighting the dual-use
risk of methods built around disinformation generation, including ours. Although our method is targeted towards counter-disinformation efforts, it could also encourage further fine-tuning of LLMs to improve
generating disinformation. Within this study, we only use models
already available on huggingface. This point needs to be taken
into account further in future work, like the question whether models
fine-tuned for generating disinformation should be released publicly
and if so, how the release can be controlled while also indicating
ethical considerations, e.g., in model cards.


\section*{Acknowledgments}

The work on this paper is performed in the scope of the projects ``VeraXtract'' (16IS24066) and ``news-polygraph'' (reference: 03RU2U151C) funded by the German Federal Ministry for Research, Technology and Aeronautics (BMFTR).

\section{Bibliographical References}\label{sec:reference}
\bibliographystyle{lrec2026-natbib}
\bibliography{custom}

\section{Language Resource References}
\label{lr:ref}
\bibliographystylelanguageresource{lrec2026-natbib}
\bibliographylanguageresource{languageresource}

\appendix

\section{Appendix}

\label{sec:appendix}

\subsection{Evaluation Details}
\subsubsection{Experimental Setup}

All experiments were run on a system with an NVIDIA H100 GPU.
The runtimes for setups based on OpenAI models reflect the inference time behind the OpenAI API.

\subsubsection{Performance per Narrative on CARDS and Statistical Analysis of Performance Correlation}
\label{app:per_narrative}
\begin{figure}[h]
    \centering
    \includegraphics[scale=0.27]{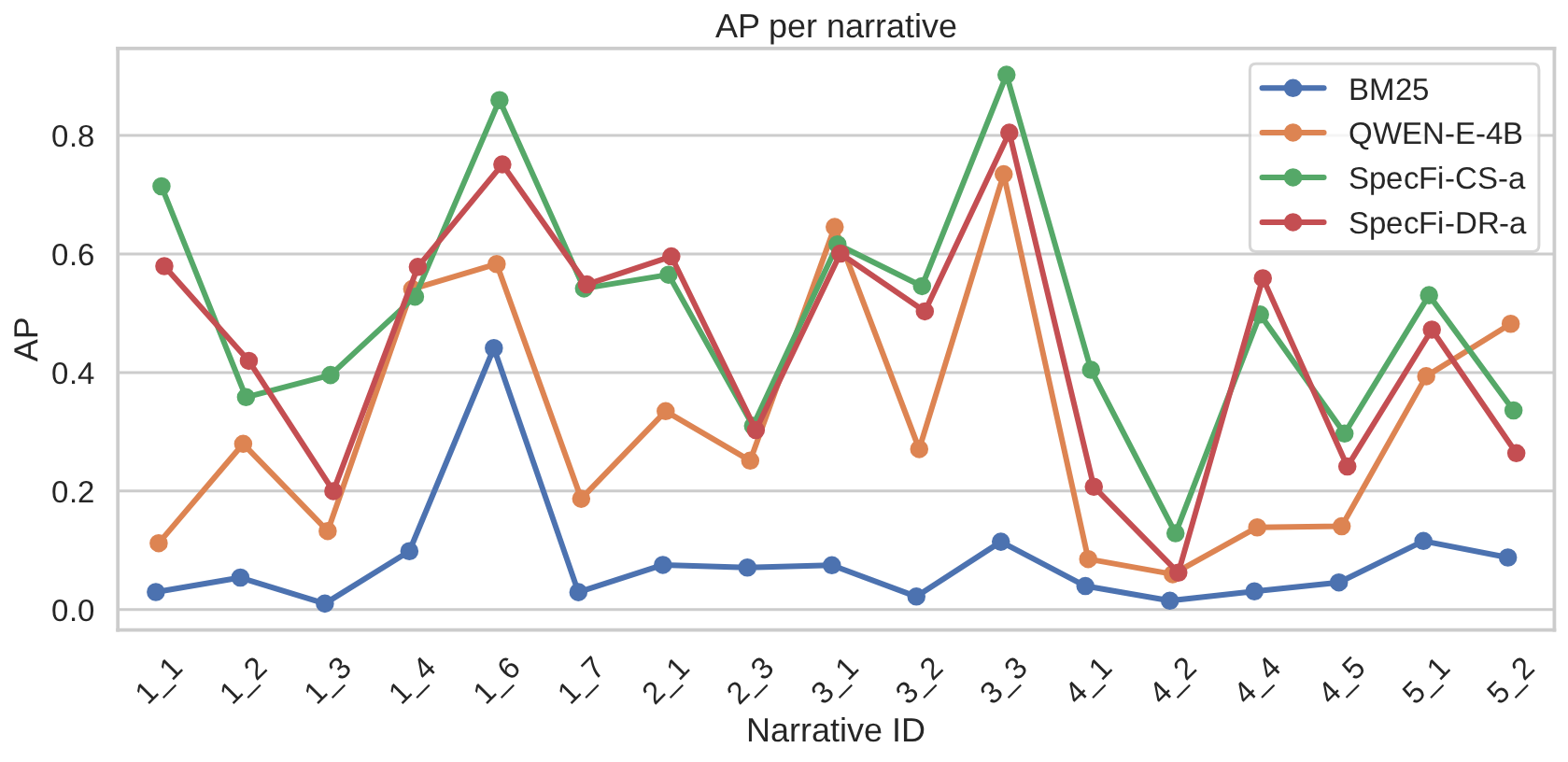}
    \caption{AP results per narrative on the CARDS dataset}
    \label{fig:ap_per_narr}
\end{figure}
As documented in Figure \ref{fig:ap_per_narr}, although the retrieval performance differs per narrative when compared over the whole CARDS dataset, there are also similarities between the results across setups. For example, the two best results for all setups are the spikes at the narrative ids 1\_6 and 3\_3, while 4\_1 and 4\_2 are some of the lowest scores for all 4 setups, notably also including SpecFi-CS.

Inspecting the community summaries retrieved as few-shot examples for each narrative reveals three distinct patterns: convergence, collapse, and drift (see Appendix~\ref{app:summaries} for the full mapping).

\paragraph{Convergence.}
For the majority of narratives, the retrieved community summary closely mirrors the expert-crafted narrative label from the CARDS taxonomy, despite the community detection operating without access to any labels. For example, the summary retrieved for narrative 1\_2 (``heading into ice age / global cooling'') states that ``the Earth may be entering a cooling cycle,'' and the summary for 5\_1 (``science is uncertain / unreliable'') expresses ``skepticism about the reliability of climate models.'' Similar convergence is observed for narratives 1\_6, 2\_1, 3\_1, and 5\_2.
We verified in the NodeRAG source code that neither filenames nor any external metadata enter the graph construction or summary generation pipeline: the LLM operates exclusively on text content extracted from the corpus.\footnote{Specifically, \texttt{Community\_summary.get\_normal\_query()} in the NodeRAG codebase aggregates the \texttt{context} field of semantic unit and attribute nodes within each Leiden partition; input filenames are stored only in a separate document tracking table and never appear in any LLM prompt.}
The convergence therefore reflects genuine bottom-up re-derivation of narrative structure from textual co-occurrence patterns in the knowledge graph.

\paragraph{Collapse.}
Where the Leiden resolution is too coarse, semantically adjacent sub-narratives merge into a single community. Narratives 4\_1 (``climate policies are harmful''), 4\_2 (``policies are ineffective''), and 4\_3 (``too difficult to solve'') all receive an identical candidate summary (``Current climate policies are criticized for being ineffective, as they rely on unrealistic targets and fail to consider political and technological realities, leading to significant market failures.''), collapsing distinct argumentative strategies into a single description. Similarly, narratives 3\_2 and 3\_3 share a summary about CO$_2$ increases being ``harmless or even beneficial.'' These narratives show correspondingly low AP for SpecFi-CS-a, suggesting that the system's few-shot examples lack the specificity needed to generate discriminative hypotheticals when sub-narratives share thematic structure. This points to community detection granularity as a key parameter for future optimization: finer-grained communities could preserve distinctions that the current Leiden resolution merges.

\paragraph{Drift.}
A third failure mode occurs when the community captures the \emph{topic} but not the \emph{stance}. The summary retrieved for narrative 4\_4 (``clean energy technology / biofuels won't work'') instead describes ``advancements in technology'' that ``can provide innovative solutions''---effectively arguing \emph{for} clean energy rather than against it. This polarity inversion likely arises because the community was dominated by texts \emph{discussing} renewable energy technology, and the LLM's summarization defaulted to the majority framing within the cluster. As a result, the generated hypotheticals are semantically opposed to the target narrative, representing a fundamentally different failure from collapse: where collapse loses granularity, drift inverts argumentative direction.

\bigskip

In Table \ref{tab:stats_perf}, we evaluate if the performance of the considered systems does indeed correlate. Several statistically significant correlations can be reported, especially within the results on CARDS and PN. For example, the comparison between BM25 and QWEN3-E-4B indicates the highest correlation with a rho value of 0.824 and a FDR-corrected p-value of 0.000 on CARDS.
Both SpecFi variants behave more independently, especially in regard to the SpecFi-CS-a setup which, for example, yields the lowest rho values when compared to BM25 results with a p-value of 0.249 on CARDS and thus with the highest value above the 0.05 significance threshold.
\begin{table}[t]
    \centering
    \resizebox{0.45\textwidth}{!}{
        \begin{tabular}{cllccccl}
              \toprule
 &  &  &  & QWEN & SpecFi & SpecFi\\
\# & Metric & Setup & BM25 & -E-4B & -DR-a & -CS-a\\
             \toprule
1& RHO & BM25 & 1.000 & 0.824 & 0.544 & 0.365 \\
 && QWEN-E-4B & 0.824 & 1.000 & 0.679 & 0.520\\
 && SpecFi-DR-a & 0.544 & 0.679 & 1.000 & 0.892\\
& & SpecFi-CS-a  & 0.365 & 0.520 & 0.892 & 1.000\\
 \cmidrule(lr){2-7}
&FDR-p & BM25 & 0.000 & \textbf{0.000} & 0.060 & 0.249\\
& & QWEN-E-4B & \textbf{0.000}& 0.000 & \textbf{0.009} & 0.065\\
& & SpecFi-DR-a & 0.060 & \textbf{0.009} & 0.000 & \textbf{0.000}\\
& & SpecFi-CS-a & 0.249 & 0.065 & \textbf{0.000} & 0.000\\
 \midrule
2&RHO  & BM25  & 1.000 & 0.786 & 0.321 & -0.071\\
& & QWEN-E-4B & 0.786 & 1.000 & 0.536 & 0.321\\
& & SpecFi-DR-a & 0.321 & 0.536 & 1.000 & 0.750\\
& & SpecFi-CS-a & -0.071 & 0.321 & 0.750 & 1.000\\
 \cmidrule(lr){2-7}
 &FDR-p & BM25 & 0.000 & 0.121 & 0.536 & 0.879\\
& & QWEN-E-4B & 0.121 & 0.000 & 0.359 & 0.536\\
& & SpecFi-DR-a & 0.536 & 0.359 & 0.000 & 0.130\\
& & SpecFi-CS-a & 0.879 & 0.536 & 0.130 & 0.000\\
 \midrule
3&RHO & BM25 & 1.000 & 0.657 & 0.489 & 0.366\\
& & QWEN-E-4B & 0.657 & 1.000 & 0.436 & 0.353\\
& & SpecFi-DR-a & 0.489 & 0.436 & 1.000 & 0.558\\
& & SpecFi-CS-a & 0.366 & 0.353 & 0.558 & 1.000\\
 \cmidrule(lr){2-7}
 &FDR-p& BM25  & 0.000 & \textbf{0.000} & \textbf{0.001} & \textbf{0.014}\\
& & QWEN-E-4B & \textbf{0.000} & 0.000 & \textbf{0.004} & \textbf{0.016}\\
& & SpecFi-DR-a & \textbf{0.001} & \textbf{0.004} & 0.000 & \textbf{0.000}\\
& & SpecFi-CS-a & \textbf{0.014} & \textbf{0.016} & \textbf{0.000} & 0.000\\
            \bottomrule
        \end{tabular}
    }
    \caption{Spearman's rho and FDR-corrected p-values for correlation analysis of performances on the MAP metric. \#1 documents the results on CARDS, \#2 on CO and \#3 on PN.}
    \label{tab:stats_perf}
\end{table}

\subsection{Community Summaries}
\label{app:summaries}
\begin{table*}[t]
\centering
\scriptsize
\setlength{\tabcolsep}{3pt}
\renewcommand{\arraystretch}{1.0}
\caption{CARDS narrative taxonomy with community summaries generated within the NodeRAG framework. Pattern categories: \emph{convergence} (summary aligns with taxonomy label), \emph{partial} (aligns at super-claim level), \emph{collapse} (identical summary shared with sibling sub-narrative), \emph{drift} (correct topic, wrong stance or focus), \emph{incoherent} (summary unrelated to narrative).}
\label{tab:summaries}

\begin{tabular}{@{}p{0.6cm}p{3.5cm}p{1.2cm}p{8.0cm}@{}}
\toprule
ID & Narrative Label & Pattern & Community Summary \\
\midrule

\multicolumn{4}{l}{\textbf{1: Global warming is not happening}} \\
\midrule

1\_1 & Ice/permafrost/snow cover isn't melting & partial
& The text explores the concept of anthropogenic global warming as a myth, questioning its validity and the narratives surrounding it. \\

1\_2 & Heading into ice age/global cooling & converg.
& There are emerging voices cautioning against the narrative of catastrophic global warming, suggesting that the Earth may be entering a cooling cycle. \\

1\_3 & Weather is cold/snowing & drift
& Severe weather events, such as unexpected snowfall, significantly affect city operations and highlight the need for preparedness in urban planning. \\

1\_4 & Climate hasn't warmed over the last decade(s) & partial
& Maps generated by climate models, used by the IPCC, are criticized as `fantasy maps' that do not accurately reflect Earth's climate history or current state. \\

1\_6 & Sea level rise is exaggerated/not accelerating & converg.
& Recent studies indicate that the rate of sea level rise has remained consistent, contradicting some climate model predictions, which raises questions about their reliability. \\

1\_7 & Extreme weather isn't increasing/has happened before & converg.
& The relationship between climate change and extreme weather events remains contentious. While some studies suggest that rising temperatures may lead to more severe weather patterns, others argue that evidence does not support a significant increase in the incidence or severity of such events. \\

\midrule
\multicolumn{4}{l}{\textbf{2: Human greenhouse gases are not causing climate change}} \\
\midrule

2\_1 & It's natural cycles/variation & converg.
& The debate over climate change often centers on the relative contributions of natural variability versus human-induced factors. While greenhouse gas emissions are acknowledged, many scientists emphasize the significant role of natural processes in shaping climate. \\

2\_3 & No evidence for greenhouse effect/CO\textsubscript{2} driving climate change & drift
& Climate change poses significant challenges to agriculture, with erratic weather patterns threatening crop yields. However, rising CO\textsubscript{2} levels may enhance plant growth, presenting a complex scenario where adaptation strategies are essential. \\

\midrule
\multicolumn{4}{l}{\textbf{3: Climate impacts/global warming is beneficial/not bad}} \\
\midrule

3\_1 & Climate sensitivity is low/negative feedbacks & converg.
& Recent studies suggest that the negative impacts of warming may not be as severe as previously believed, indicating that CO\textsubscript{2} climate sensitivity is significantly lower than earlier estimates. \\

3\_2 & Species/plants/reefs benefiting from climate change & converg.
& Some argue that the effects of CO\textsubscript{2} increases and slight global warming may be harmless or even beneficial, challenging alarmist narratives about climate change. \\

3\_3 & CO\textsubscript{2} is beneficial/plant food & collapse
& \emph{(Same as 3\_2)} \\

\midrule
\multicolumn{4}{l}{\textbf{4: Climate solutions won't work}} \\
\midrule

4\_1 & Climate policies are harmful & converg.
& Current climate policies are criticized for being ineffective, as they rely on unrealistic targets and fail to consider political and technological realities, leading to significant market failures. \\

4\_2 & Climate policies are ineffective/flawed & collapse
& \emph{(Same as 4\_1)} \\

4\_4 & Clean energy/biofuels won't work & drift
& The text explores how advancements in technology can provide innovative solutions to combat climate change, including renewable energy sources and carbon capture methods. \\

4\_5 & People need energy (fossil fuels/nuclear) & incoherent
& The presence of the Wolverine at the convention stage highlights how external factors can influence the performance and effectiveness of energy panels, suggesting a need for adaptive strategies in energy management. \\

\midrule
\multicolumn{4}{l}{\textbf{5: Climate movement/science is unreliable}} \\
\midrule

5\_1 & Science is uncertain/unsound/unreliable & converg.
& Many scientists express skepticism about the reliability of climate models, which have been criticized for failing to accurately predict temperature changes and for being overly reliant on theoretical calculations. \\

5\_2 & Movement is alarmist/political/biased & converg.
& The authors argue that the climate alarmism movement is losing credibility, with outdated predictions and ideological biases undermining its claims. \\

\bottomrule
\end{tabular}
\end{table*}

\subsection{Prompts Collection}
 
\subsubsection{Embedding Models}
 
The default model instruction prompt is:
\begin{lstlisting}[style=prompt]
Instruct: Given a web search query, retrieve relevant passages that answer the query
Query: {query}
\end{lstlisting}
 
The following prompt was used to enhance the retrieval results:
\begin{lstlisting}[style=prompt]
Instruct: Given a narrative description as a query, retrieve passages that serve this narrative; can be entailed from the narrative; can be aligned logically with the narrative
Query: {query}
\end{lstlisting}
 
Within our HyDE-based setup, we use the following instruction prompt for embedding models:
\begin{lstlisting}[style=prompt]
Instruct: Given a text as a query retrieve relevant passages that align with narratives similar to the query
Query: {query}
\end{lstlisting}

\subsubsection{Generative Models}
 
Our system prompt:
\begin{lstlisting}[style=prompt]
You are a disinformation investigator. Your first step is to generate short disinformation texts that sound like actual ones. You get a disinformation narrative and return a disinformation text that aligns with that narrative. Return only 1 single text!
\end{lstlisting}
 
Each system prompt was followed by a user prompt:
\begin{lstlisting}[style=prompt]
You are a disinformation investigator. Given a disinformation narrative, generate a short, realistic text (such as a news excerpt, blog post, or social media post) that supports or aligns with that narrative. The text should sound plausible and could be found in the wild.
 
Here are some examples: {examples}
 
Narrative: {query}
Text:
\end{lstlisting}
 
\subsection{Narrative Taxonomies Transformed to Search Queries}
\label{sec:queries}
 
All datasets included predefined narrative taxonomies. We constructed narrative descriptions from the narrative labels with the following strategies:
 
\subsubsection{CARDS}
The queries were constructed by following the CARDS taxonomy and the corresponding id labeling with the dataset and concatenating ``claim'' (=subnarrative) and ``superclaim'' (=narrative) descriptions from the code book.
 
\begin{lstlisting}[style=prompt]
"1_1": "Global warming is not happening. Ice/permafrost/snow cover isn't melting",
"1_2": "Global warming is not happening. We're heading into an ice age/global cooling",
"1_3": "Global warming is not happening. Weather is cold/snowing",
...
"5_3": "Climate movement/science is unreliable. Climate change (science or policy) is a conspiracy (deception)",
\end{lstlisting}
 
Resulting in predefined 27 narratives in the taxonomy, out of which 17 can be found in the dataset.
 
\subsubsection{Climate Obstruction}
 
Constructed by using the narrative ids from the dataset and descriptions of the narratives provided in the supplemental material of the paper.
 
\begin{lstlisting}[style=prompt]
"CA": "Community & Resilience. Emphasizes how the oil and gas sector contributes to local and national economies through tax revenues, charitable efforts, and support for local businesses",
"CB": "Community & Resilience. Focuses on the creation and sustainability of jobs by the oil and gas industry."
...
"SA": "Patriotic Energy mix. Stresses how domestic oil and gas production benefits the nation, including energy independence, energy leadership, and the idea of supporting American energy"
\end{lstlisting}
 
Resulting in 7 narratives, out of which all can be found in the dataset.
 
\subsubsection{PolyNarrative}
Similar to CARDS, the queries were constructed by following the PolyNarrative (PN) taxonomy and the corresponding id labeling with the dataset.
 
\begin{lstlisting}[style=prompt]
"1_1": "Blaming the war on others rather than the invader: Ukraine is the aggressor",
"1_2": "Blaming the war on others rather than the invader: The West are the aggressors",
...
"21_2": "Green policies are geopolitical instruments: Green activities are a form of neo-colonialism"
\end{lstlisting}
 
Resulting in predefined 88 narratives in the taxonomy, out of which 51 can be found in the dataset.

\end{document}